\documentclass[journal,10pt,final]{IEEEtran}
\usepackage{xcolor}
\usepackage{threeparttable}
\usepackage{booktabs}
\usepackage{multirow}

\usepackage{float}
\usepackage{diagbox}

\usepackage{cite}

\ifCLASSINFOpdf
  \usepackage[pdftex]{graphicx}
  \graphicspath{{figs/}}
  \DeclareGraphicsExtensions{.pdf}
\else
\fi
\usepackage{amsmath}
\ifCLASSOPTIONcompsoc
\usepackage[caption=false,font=normalsize,labelfon
t=sf,textfont=sf]{subfig}
\else
\usepackage[caption=false,font=footnotesize]{subfi
g}
\fi

\usepackage{algorithm}
\usepackage{algorithmicx}
\usepackage{algpseudocode}

\usepackage{array}
\begin{document}
%
\title{Lifelong Vehicle Trajectory Prediction Framework Based on Generative Replay}


\author{\IEEEauthorblockN{Peng Bao\IEEEauthorrefmark{1},
Zonghai Chen\IEEEauthorrefmark{1},
Jikai Wang\IEEEauthorrefmark{1}, 
Deyun Dai\IEEEauthorrefmark{1} and
Hao Zhao\IEEEauthorrefmark{1}}%

\IEEEauthorblockA{\IEEEauthorrefmark{1}Department of Automation, University of Science and Technology of China, Hefei, PR China}
\thanks{Corresponding author: Z. Chen (email: chenzh@ustc.edu.cn).}}

%



\IEEEtitleabstractindextext{%
\begin{abstract}
Accurate trajectory prediction of vehicles is essential for reliable autonomous driving. To maintain consistent performance as a vehicle driving around different cities, it is crucial to adapt to changing traffic circumstances and achieve lifelong trajectory prediction model. To realize it, catastrophic forgetting is a main problem to be addressed. In this paper, a divergence measurement method based on conditional Kullback-Leibler divergence is proposed first to evaluate spatiotemporal dependency difference among varied driving circumstances. Then based on generative replay, a novel lifelong vehicle trajectory prediction framework is developed. The framework consists of a conditional generation model and a vehicle trajectory prediction model. The conditional generation model is a generative adversarial network conditioned on position configuration of vehicles. After learning and merging trajectory distribution of vehicles across different cities, the generation model replays trajectories with prior samplings as inputs, which alleviates catastrophic forgetting. The vehicle trajectory prediction model is trained by the replayed trajectories and achieves consistent prediction performance on visited cities. A lifelong experiment setup is established on four open datasets including five tasks. Spatiotemporal dependency divergence is calculated for different tasks. Even though these divergence, the proposed framework exhibits lifelong learning ability and achieves consistent performance on all tasks.    
\end{abstract}

}

\maketitle

\IEEEdisplaynontitleabstractindextext

%
\IEEEpeerreviewmaketitle

\section{Introduction}
%
%
%
%
\IEEEPARstart{I}{n} autonomous driving, an ability to predict surrounding vehicles' future trajectories accurately is a key to make appropriate decision. Therefore, many research works contribute on vehicle kinematics and interactive modelling\cite{2017Predicting}. Recent works also adopt data-driven approaches\cite{2019Interactive,li2019interaction,li2019grip,chandra2019traphic,dulian2021multi,bahari2021injecting,zyner2019naturalistic}. Benefitted from these ingenious works, precision of vehicle trajectory prediction on various open datasets has been promoted significantly. 

However, in real application, an intelligent vehicle equipped with an autonomous driving system is supposed to visit varied road sections, cities or even countries. To guide the vehicle safely, the system is required to adapt to heterogeneous distribution of surrounding vehicles' motion and interaction pattern and predict their future trajectories accurately. For this purpose, the system needs to learn new knowledge about emerging traffic environments continuously without forgetting old ones. In addition, with limited storage resource, the system can not afford to store large amount of trajectory data. Coutinuous learning with limited storage resource to achieve good performance in all processed tasks is also called lifelong learning. Unfortunately, most existing vehicle trajectory prediction models are trained and tested specifically for each dataset, which fails to accomodate to other different datasets if applied directly. 

A crucial problem impeding lifelong learning is catastrophic forgetting. In different traffic circumstances, interaction pattern among vehicles varies, which can be interpreted as spatiotemporal dependency divergence. The divergence makes a prediction model trained in old circumstances performs poorly in a new different circumstance. If trained in the new circumstances, the model would fit spatiotemporal dependency in the new one while forget that in the old circumstances, which makes it perform poorly in the old circumstances in turn. In summary, to perform consistently in all circumstances, the prediction model is required to learn new knowledges while not forgetting old ones.

Therefore, to measure spatiotemporal dependency divergence that arises from varied traffic circumstances, Mixture Density Networks (MDNs) are introduced to estimate conditional probability density function (CPDF) and conditional Kullback-Leibler divergence (CKLD) is computed through Monte-Carlo (MC) sampling. Then to realize lifelong trajectory prediction, a new framework based on conditional generative replay is proposed. The framework consists of two models, a generative memory and a task solver. The generative memory is designed to address catastrophic forgetting, a key challenge in lifelong learning. Through adversarial learning and distributions merging across different traffic circumstances, the generative memory replays trajectories conditioned on spatial configuration of vehicles. The replayed trajectories is used as an input to train a trajectory prediction model. As the generative memory is capable of generating trajectories within the same distribution of all recorded data, the prediction model achieves consistent performance on all processed tasks. 

In summary, our contributions include:
\begin{itemize} 
\item An innovative spatiotemporal dependency divergence measurement method is developed for two trajectory datasets.
\item A novel research topic is revealed. Lifelong vehicle trajectory prediction is first proposed to promote adaptation and performance on heterogeneous tasks for autunomous driving system.  
\item Based on generative replay, an initiate lifelong learning framework is introduced for vehicle trajectory prediction.  
\item Experimental results on four different datasets have demonstrated that the proposed framework can achieve lifelong learning and realize satisfactory prediction performance on all processed tasks. 
\end{itemize}

\section{Related work}

\textbf{Vehicle trajectory prediction} is a longstanding research topic and is becoming more important as the development of autonomous driving. Kinematics models\cite{ammoun2009real,lytrivis2008cooperative} are established to predict surrounding vehicles' future trajectories, which performs well in short horizon. For a longer prediction horizon over 3 seconds, vehicles' intentions are important and have to be clarified first. Prototype classification\cite{tran2014online} and intention recognition\cite{aoude2011behavior} are two research directions. Through collecting and clustering of tremendous trajectories in advance, prototype classification methods match online history trajectories with that in database. Multimodel future trajectories can be generated once matched. Intention recognition methods classify drivers' potential maneuvers into limited classes. Combining heuristic information including road topology, traffic signal and vehicle turn signal, classification models such as Support Vector Machines (SVMs)\cite{kumar2013learning} and Hidden Markov Models (HMMs)\cite{streubel2014prediction} are introduced to recognize surrounding drivers' intentions. However, drivers' behaviours are affected by various factors and highly personalized\cite{xu2015establishing}, which makes it hard to be recognized accurately in real application. Besides, a vehicle's trajectory is not determined by its driver's intention only but affected interactively. 

To make an accurate prediction, it's necessary to consider vehicles' motion as a dynamic interactive system and model interactive impact among vehicles. Pairwise interactive modelling is of high complexity. Deo Nachiket\cite{deo2018would} simplified interactive factor as a cost function that penalizes vehicle collision to filter out collide future trajectory pairs. Data-driven approaches are another practical methods and have been digged extensively. In these approaches, vehicles' sequential features are learned through Long-Short Term Memory Networks (LSTMs). To learn interactive factor, Convolutional Neural Networks (CNNs) or max-pooling module is introduced. Deo\cite{deo2018convolutional} divided certain surrounding area into grid cells and use CNNs to model spatial relationship among vehicles. Li\cite{li2019grip} builded an adjacent matrix of surrounding vehicles where each elements represents pairwise proximity. A CNN is utilized to learn interactive factor. Messaoud\cite{messaoud2019relational} discretized traffic environment into 3D grid cells and a Relational Recurrent Neural Network (RRNN) is used to predict future trajectory. Gupta\cite{gupta2018social} utilized a pooling module to extract dense interactive features among vehicles. More recent works urge to merge data-driven and knowledge-driven approaches into a unified neural network\cite{dulian2021multi,bahari2021injecting}.

However, although prediction precision is improved continually, generalization and adaptation of proposed models still remain an open problem. Majority of research works validate proposed methods on one open dataset only. Some works validate on more datasets but train and test models for each dataset individually, which is not consistent with real application. Pushing forward to real application, a novel prediction model is urged to be proposed to cope with various traffic circumstances. Two questions arise naturally. How to measure difference among heterogeneous circumstances? And how to fulfill consistent vehicle trajectory prediction performance on emerging circumstances and visited ones, which is also called lifelong learning?

\textbf{Divergence measurement of heterogeneous traffic circumstances} is an open problem. To measure trajectory similarity, various methods are proposed, such as euclidean distance (ED)\cite{sanderson1980pattern}, dynamic time warping (DTW)\cite{muller2007dynamic}, longest common subsequence (LCSS)\cite{robinson1990temporal}, merge distance (MD)\cite{ismail2015new}, and spatiotemporal locality in-between polylines (STLIP) distance \cite{pelekis2007similarity}, etc. Su\cite{su2020survey} made a survey on 15 widely used trajectory distance measures in the literature. It can be deduced that ED is suitable for vehicle trajectories distance measurement that have the same total length and sample frequency. However, these similarity measurement methods consider one trajectory with another trajectory each time, while we aim to measure differences of traffic circumstances where dynamic number of trajectories are presented. In different traffic circumstances, vehicles' interaction pattern changes and the way affecting future motion varies. In other words, spatiotemporal dependency between future and past motion differs, which is essentially a conditional probability density function (CPDF) alteration problem. Therefore, a more reasonable method is to estimate distance between two unknown CPDF with empirical samples only.

\textbf{Estimation distance of two unknown CPDF with samples only} is challenging. As a commonly used probability divergence measurement method, KL divergence can not work without analytic CPDF. The Donsker-Varadhan variational formula \cite{https://doi.org/10.1002/cpa.3160360204,belghazi2018mutual} can be utilized to estimate CKLD empirically. However, it suffers from convergence problem for large divergence between two CPDFs, which is usually the case in real data. $K$ nearest neighbor\cite{wang2006nearest} is another approch to approximate CKLD. It requires distance calculation of condition data for two datasets, which is not commited for traffic circumstances with dynamic number of vehicles. As a conditional extension of traditional Maximum Mean Discrepancy (MMD)\cite{gretton2006kernel}, conditional MMD (CMMD)\cite{song2013kernel} is proposed to measure distance between two CPDFs. Similar with MMD, CMMD measures embedding probabilities distance in reproducing kernel Hilbert space (RKHS). In MMD, a PDF is mapped into a point in RKHS, while in CMMD a CPDF is a family of points with different conditions. Therefore, CMMD is averaged for distances with different condition, which implies samples conditioned on a same condition. For two traffic circumstances in our work, condition data can not guaranteed to be the same. In fact, CMMD is usually used as training loss function\cite{ren2016conditional,ren2019learning} of neural networks where predicted and real value can be obtained on the same condition. Another method to measure PDF distance empirically is optimal transport. Esteban\cite{tabak2021data} proposes a data-driven conditional optimal transport (COT) method. The COT represents empirical CPDF distance computation as a optimal transport problem constrained by CPDF alignment. CKLD is utilized to interpret the constraint and then converted into KL divergence between joint distributions through chain rule\cite{cover1999elements}. By using Donsker-Varadhan variational formula and Lagrange multiplier, the constrained COT is relaxed into a minimax optimization problem that can be optimized empirically. However, the COT prones to local minimum and the minimax game is hard to converge. Moreover, for high dimension data as in our work, it is difficult to identify local minimum.

\textbf{Lifelong learning} aims to solve a series of tasks incrementally\cite{delange2021continual}. When addressing a new task, small amount or none data of old tasks are stored. After the final task is presented, all tasks should be solved by one task model with good performance. Key to lifelong learning is avoiding catastrophic forgetting of old tasks' knowledge when updating the task model to solve a new task. From an aspect of model training, approches to mitigate catastrophic forgetting are classified into three categories, architectural, regularization and rehearsal strategies. Architectural strategies train different models or subnetworks for incoming new tasks. A selector is used to choose an appropriate model or subnework for a task. Typical research works includes Progressive Neural Network (PNN)\cite{rusu2016progressive}, Incremental Learning through Deep Adaptation (DAN)\cite{rosenfeld2018incremental}, Copy Weight with Re-init (CWR)\cite{lomonaco2017core50}, etc. These methods preserve performances of old tasks while confilcting with storage limitation of lifelong learning. Regularization strategies extend loss functions with additional term to retain performances of old tasks. Learning without Forgetting (LwF)\cite{li2017learning} and Elastic Weight Consolidation (EWC)\cite{kirkpatrick2017overcoming} are two representative methods. LwF proposes to use outputs of old models as soft targets to substitute data of old tasks, which is reported to suffer a buildup drop in old tasks' performance as the task sequence grows longer\cite{aljundi2017expert}. EwC evaluates importance of parameters for old tasks and adds a penalty to changes when training on new tasks, which pays more attention to preserving the knowledge on old tasks but prevents the model from achieving competitive performance on new tasks\cite{yao2019adversarial}. Rehearsal strategies generally use an external memory to store part of old data\cite{hou2018lifelong,rebuffi2017icarl} or patterns\cite{lopez2017gradient}. As storage is limited and Generative Adversarial Networks (GANs) develope, generative replay\cite{shin2017continual} is proposed as a memory of previous data and its feasibility has been validate on several works\cite{su2019generative,lesort2019generative,lesort2019marginal,liu2020generative,li2020incremental}. Although quality of the generation model is a bottleneck, many works\cite{van2018generative, kamra2017deep,wu2018memory,liu2020generative,su2019generative, xiang2019incremental} have proved that an elaborately designed generation model practically outperforms mainstream lifelong methods such as EWC, LwF, MAS\cite{aljundi2018memory}, PathNet\cite{fernando2017pathnet}, and iCaRL\cite{rebuffi2017icarl} \emph{et al}.

In this paper, CPDF distance between two traffic circumstances are calculated first to reveal spatiotemporal dependency divergence. Then a generative replay based lifelong trajectory prediction framework is proposed to enhance generalization and adaptation over different traffic environment. As a key of alleviate catastrophic forgetting, a generation model is realized through a novel conditional GAN (CGAN), which is called Recurrent Regression GAN (R2GAN). Through merging different generation models trained on different tasks, the generation model finally learns all knowledge involved in processed tasks. Eventually, a trajectory prediction model trained on generated data performs well on all tasks. A task chain including five tasks that stem from four open datasets is used as lifelong setup. Experiments on the task chain demonstrate effectiveness of proposed framework.   

The rest of this paper is organized as follows. In section 3, a mathematic formulation for lifelong trajectory prediction is addressed. Divergence between two traffic circumstances is measured first in section 4. Then generative replay based lifelong prediction framework is introduced in section 5 in detail. In section 6, evaluation experiments are performed to evaluate quantitatively. Finally, conclusion and future work are introduced in section 6.

\section{Problem Formulation}
Formally, lifelong vehicle trajectory prediction is characterized by a set of tasks $D=\{d_1,d_2,\ldots,d_n\}$ to be learned by a parameterized model. In this work, with unsupervised learning nature of trajcetory prediction, task data $d_i \in D$ have training samples ${X_{1:t}^i}$ where $t=t_h+t_f$. Target vehicle and surrounding vehicles' trajectories lasting for $t_h$ are regarded as history information to train the parameterized model to predict future $t_f$ trajectory of the target vehicle. In a traffic circumstance involving $n_d$ vehicles, spatiotemporal dependency is formulated as a CPDF $p\left(Y|X\right)$ where $Y$ represents future $t_f$ trajectory of the target vehicle and $X$ represents history $t_h$ trajectories of all vehicles. Samples are drawn \emph{i.i.d} from an unknown distribution $X_{1:t}^{i}\in P_{d_i}$ associcated with task $d_i$. Distribution $P_{d_i}$ can be different from each other for different $i \in\{1,2, \ldots, n\}$. In lifelong learning, task data $d_i$ are observed sequentially and when the next data $d_{i+1}$ arrive, data ${d_i}$ are abandoned completely of only kept partly in a limited storage. Ultimately, the prediction model can predict accurately in all $n$ tasks after observing all task data.   

\section{Divergence measurement of different traffic circumstances}
As an effective divergence measuremnt method, KL divergence is extended to CKLD to measure spatiotemporal dependency difference of two traffic circumstances, which is formulated as
\begin{equation}
\begin{aligned}
\label{CKLD_def}
&CKLD\left(p_1\left(Y\left|X\right. \right)||p_2\left(Y\left|X\right. \right)\right)\\&=\int p_{1}(X) \int \log \left(\frac{p_{1}(Y \left| X\right.)}{p_{2}(Y \left|X\right.)}\right) p_{1}(Y \left|X\right.)dYdX.
\end{aligned}
\end{equation}
The CKLD can not be computed without analytic formulation of $p\left(Y \left|X\right.\right)$. Therefore, parameters of GMMs are estimated by a MDN to approximate $p\left(Y \left|X\right.\right)$ first and then MC sampling can be performed to calculate CKLD.
\subsection{Dimension normalization for dynamic traffic circumstances}
In a dynamic traffic circumstances involving $n_d$ vehicles, condition $X$ should be represented as $X=(x_{1:t_h}^{1},x_{1:t_h}^{2},\dots,x_{1:t_h}^{n_d})$ where $x_{1:t_h}^{i}$ represents sequential coordinate of vehicle $i$ that lasts $t_h$, which possesses dynamic dimension. To facilitate model learning, a fixed dimension is preferred.
Notice that target vehicle's future motion is affected by limited number of neighboring vehicles, it is reasonable to consider $n_v$ closest vehicles only, which is also a common practice in trajectory prediciton research\cite{li2019grip,lee2019joint}. To represent interactive relationship between considered vehicles, a Laplacian matrix is calculated. Being different from usual 3D case\cite{chandra2020forecasting}, a 2D Laplacian matrix is calculated through weighting on time dimension. Then eigen vectors corresponding to the biggest $k$ eigen values are concatenated with target vehicle's history trajectory, which forms a condition vector $X=\left(x_{1:t_h}^{e},v_{1},\dots,v_{k}\right)$ with fixed dimension $d_X=2t_h+kn_v$ where the superscript $e$ represents the target vehicle. The Laplacian matrix is calculated through
\begin{equation}
\begin{aligned}
\label{laplacian}
&L=D-A,\\
&A=\left(a_{ij}\right)_{n_v\times n_v},\\
&D=\left(d_{ij}\right)_{n_v\times n_v},\\
&a_{ij} = \exp \left(-\sum\nolimits_{k=1}^{t_h} w_{k}d\left(x_{k}^{i},x_{k}^{j}\right) / \sum\nolimits_{k=1}^{t_h} w_{k}\right),\\
&w_{k} = \lambda^{t_h-k}, k=1, \ldots, t_h,\\
&d_{i j}=\left\{\begin{array}{c}
\sum_{j=1}^{n_v} a_{i j}, i=j \\
0, i \neq j
\end{array}\right.,
\end{aligned}
\end{equation}
where $d\left(x_{k}^{i},x_{k}^{j}\right)$ is ED between vehicle $i$ and $j$ at time $k$ and $\lambda$ is a decay parameter.
\subsection{Estimation on GMMs based on a MDN}
To calculate CKLD between two CPDFs, GMMs are introduced to approximate CPDF as $p(Y\left|\right. X)=\sum_{i=1}^{m} \alpha_{i}(X) \phi_{i}(Y \left|\right. X)$, where $m$ is number of gaussian distribution hypothesis and $\phi_{i}(Y \left|\right. {X})=\exp \left(-\frac{\left\|Y-\mu_{i}(X)\right\|^{2}}{2 \sigma_{i}(X)^{2}}\right)/{(2 \pi)^{d_X / 2} \sigma_{i}(X)^{d_X}} $. For $i=1,\dots,m$, mixing coefficient $\alpha_{i}(X)$, mean $\mu_{i}(X)$, and variance $\sigma_{i}(X)$ are estimated through MDN\cite{bishop1994mixture,rothfuss2019conditional}. As shown in \figurename{1}, a Multi-Layer Perceptron (MLP) is applied for input $X$ to obtain a feature encoding $Z$. Then three seperative Fully Connected (FC) layers are utilized to calculate parameters of GMMs. To enforce $\sum_{i=1}^{m} \alpha_{i}(x)=1$, a softmax function is applied $\alpha_{i}(X)={\exp \left(FC(Z)_{i}\right)}/{\sum_{j=1}^{m} \exp \left(FC\left(Z)_{j}\right)\right.}$, where $FC(\bullet)$ represents a FC layer and the subscript $i$ and $j$ represents vector component. Means are unconstrained. Variances $\sigma_{i}(X)$ should be positive. A softplus function is applied hence $\sigma_{i}(X)=\log \left(1+\exp \left(FC(Z)_{i}\right)\right)$. Training loss function for MDN is $L_{mdn}=-log\left(Y\left|X\right.\right).$
\begin{figure}[!t]
	\centering
	\includegraphics[width=3.0in]{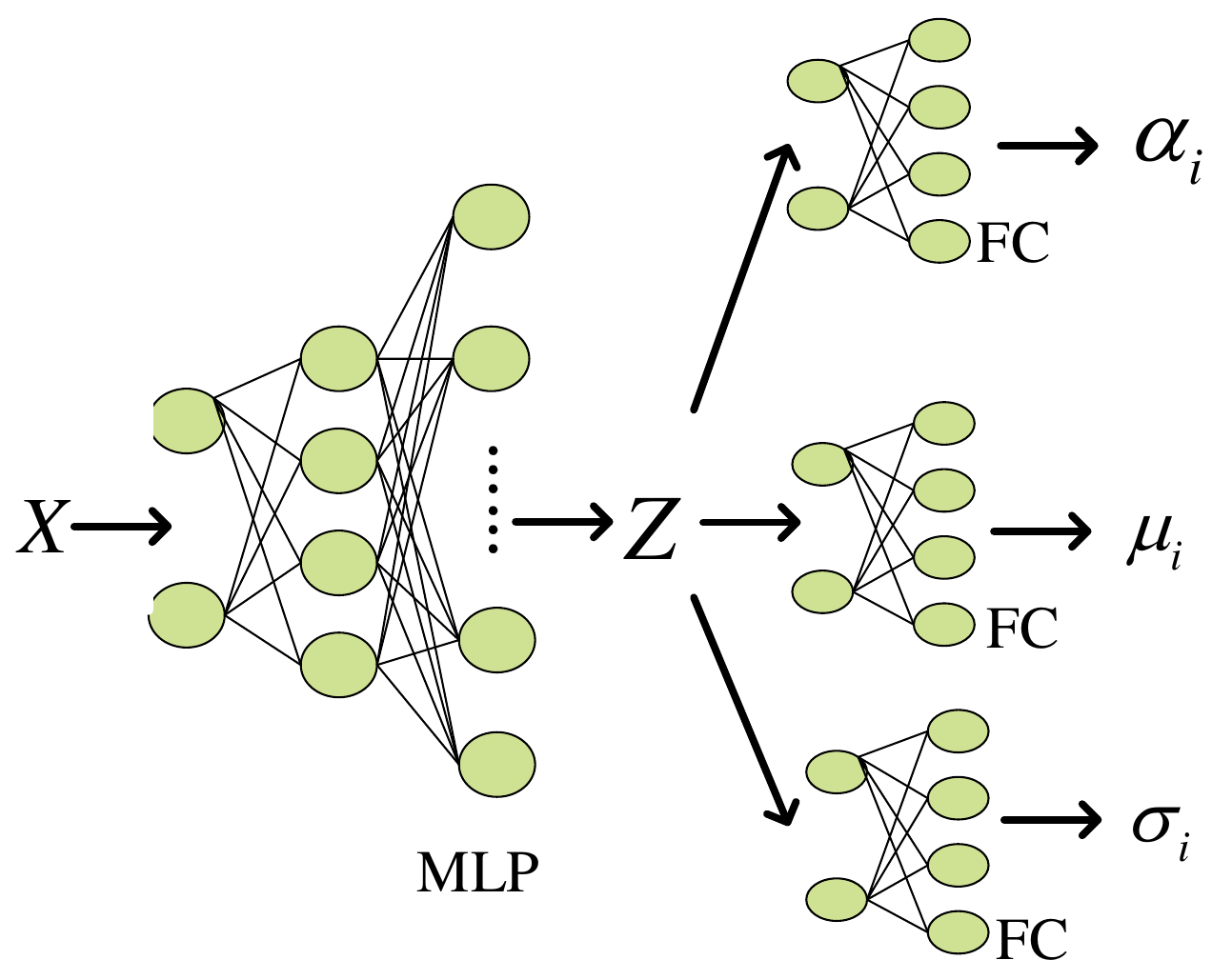}		
	\caption{MDN architecture. }
\end{figure}
\subsection{Calculation of CKLD through Monte-Carlo sampling}
After GMMs are estimated for each condition $X$, CKLD can be computed. As in (\ref{CKLD_def}), for every sample condition $X_i,i=1,\dots,n_1$ on $p_1\left(X\right)$, KLD can be calculated as 
\begin{equation}
\begin{aligned}
&KLD\left(p_1\left(Y\left|X_i\right. \right)||p_2\left(Y\left|X_i\right. \right)\right)\\&=\int \log \left(\frac{p_{1}(Y \left| X_i\right.)}{p_{2}(Y \left|X_i\right.)}\right) p_{1}(Y \left|X_i\right.)dY.
\end{aligned}
\end{equation}
Although KL divergence between two GMMs is not analytically attractable, some techniques are developed to estimate effectively. Hershey\cite{hershey2007approximating} compared 5 methods and concluded that MC sampling reaches clearly the best accuracy. Suppose samples $Y_j,j=1,\dots,n_{mc}$ are sampled from $p_1\left(Y\left|\right.X_i\right)$, then $CKLD$ can be calculated as 
\begin{equation}
\begin{aligned}
\label{CKLD_calculation}
&KLD\left(p_{1}(Y \left|X_i\right.) \| p_{2}(Y \left|X_i\right.)\right)\\&=\sum\nolimits_{j=1}^{n_{m c}}\left(\log p_{1}\left(Y_{j} \left|X_{i}\right. \right)-\log p_{2}\left(Y_{j} \left|X_{i}\right.\right)\right) / n_{mc},\\
&CKLD\left(p_{1}(Y \left|X\right.) \| p_{2}(Y \left|X\right.)\right)\\&=\sum\nolimits_{i=1}^{n_{1}}KLD\left(p_{1}(Y \left|X_i\right.) \| p_{2}(Y \left|X_i\right.)\right) / n_{1}.
\end{aligned}
\end{equation}
The complete $CKLD$ computation flow is summarized in Algorithm 1.
\begin{figure}[!t]
  \begin{algorithm}[H]
    \caption{CKLD between two traffic circumstances}
    \label{alg1}
    \begin{algorithmic}
      \Require Sample pairs $\left(X_{1i},Y_{1i}\right)\sim p_1(X,Y),i=1,\dots,n_1$ and $\left(X_{2i},Y_{2i}\right)\sim p_2(X,Y),i=1,\dots,n_2$.
      \Ensure $CKLD\left(p_{1}(Y \left|X\right.) \| p_{2}(Y \left|X\right.)\right)$.
      \State Calculate Laplacian matrix according to (\ref{laplacian}) and normalize condition to uniform dimension.
      \For{$i=1,\dots,n_1$}
        \State Fit a MDN with $\left(X_{1i},Y_{1i}\right)$ and loss function $L_{mdn}$. 
      \EndFor
      \For{$i=1,\dots,n_2$}
        \State Fit a MDN with $\left(X_{2i},Y_{2i}\right)$ and loss function $L_{mdn}$. 
      \EndFor
      \State $CKLD \gets 0$
      \For{$i=1,\dots,n_1$}
        \State Sampling $Y_j \sim p_{1}(Y \left|X_i\right.)$, $j=1,\dots,n_{mc}$
        \State Calculate KLD according to (\ref{CKLD_calculation}).
        \State $CKLD \gets CKLD+KLD\left(p_{1}(Y \left|X_i\right.) \| p_{2}(Y \left|X_i\right.)\right)$. 
      \EndFor
      \State $CKLD \gets CKLD/n_1 $.
      \State \Return CKLD
    \end{algorithmic}
  \end{algorithm}
\end{figure}
\section{Generative Replay based lifelong trajcetory prediction}
Based on our previous research on trajectory generation\cite{BAO2022370}, a novel trajectory generation model trained by the standard GAN\cite{goodfellow2014generative} loss is proposed to memorize data distribution of tasks. With a vehicle trajectory prediction model, a lifelong vehicle trajectory prediction framework is realized.

\subsection{Generator Conditioned on Relative Position Configuration}
In a vehicle trajectory prediction task, we need to predict a target vehicle's future $t_f$ trajectory according to its $t_h$ history trajectory and its neighboring vehicles' histories. To facilitate generation process, a full prediction scenario is required to be generated that consists of target vehicle's and its neighboring vehicles' trajectories lasting for $t=t_h+t_f$ horizon.

Being different from traditional GANs that model generation procedure as mapping from a prior probability distribution to a target one, we are inspired by Quant GAN\cite{wiese2020quant} and model prediction scenario generation as mapping between stochastic processes. Gaussian process with RBF kernel is selected as prior stochastic process. For single vehicle, $m$ Gaussian process samplings are obtained, which constitutes $\left(N_{1:t}^{1}, N_{1: t}^{2}, \ldots, N_{1: t}^{m n}\right)$ for total $n$ interactive vehicles. Conditional GANs are easier to train and make generated samples more controllable. Therefore, multiple vehicles' spatial configuration condition $\left(C_{1}, C_{2}, \ldots, C_{n}\right)$ is utilized as conditional inputs for our generator, where $C_{i}$ represents condition for vehicle $i$. Spatial configuration condition $\left(C_{1}, C_{2}, \ldots, C_{n}\right)$  and input sampling $\left(N_{1: t}^{1}, N_{1: t}^{2}, \ldots, N_{1: t}^{m n}\right)$  are encoded by two MLPs individually first. To map sequential feature of Gaussian process into target data, a bidirectional GRU is utilized, where initial hidden states $h_{gf0}$ and $h_{gb0}$ are set by encoded position condition\cite{karpathy2015deep,vinyals2015show}. In the bidirectional GRU, forward code ${E_f}_{1: t}^{j}$ of agent $j$ and backward code ${E_b}_{1: t}^{j}$ are averaged to form sequential   code $E_{1: t}^{j}$. A MLP is attached later to encode spatial relations. A fully connected (FC) layer is used later with tanh activation function to output spatiotemporal data $\left(\tilde{X}_{1: t}^{1},\tilde{X}_{1: t}^{2}, \ldots, \tilde{X}_{1: t}^{n}\right)$ of $n$ agents. Complete framework is shown in \figurename{2}.

Let $G\left(\bullet\right)$ represents a GRU and $M\left(\bullet\right)$ a MLP unit, the generation procedure can be formulated as:
\begin{equation}
\label{eqn_1}
\begin{aligned}
{{E_f}_{1:t}^{j}}&={G_{gf}\left(\begin{array}{l}M_{g0}\left(N_{1:t}^{mj-m}, N_{1:t}^{mj-m+1}, \ldots, N_{1:t}^{mj}\right) \\ h_{gf0}=M_{g 1}\left(C_{j}\right)\end{array}\right)},\\
{{E_b}_{1: t}^{j}}&={G_{g b}\left(\begin{array}{l}M_{g 0}\left(N_{1: t}^{m j-m}, N_{1: t}^{m j-m+1}, \ldots, N_{1: t}^{m j}\right) \\ h_{g b 0}=M_{g 2}\left(C_{j}\right)\end{array}\right)},\\
{E_{1:t}^{j}}&={M_{g 3}\left(\left({E_f}_{1: t}^{j}+{E_b}_{1: t}^{j}\right) / 2\right)}, \\
E_{1:t}^{i,j}&=E_{1:t}^{i}-E_{1:t}^{j},\\
\tilde{X}_{1:t}^{j}&=M_{g5}\left(E_{1:t}^{j},\left(\sum\nolimits_{i=1,i\neq j}^{n}M_{g4}\left(E_{1:t}^{i,j}\right)\right)/(n-1)\right),
\end{aligned}
\end{equation}
for agent $i,j=1,2, \ldots, n.$

\begin{figure*}[!t]
	\centering
	\includegraphics[width=6.0in]{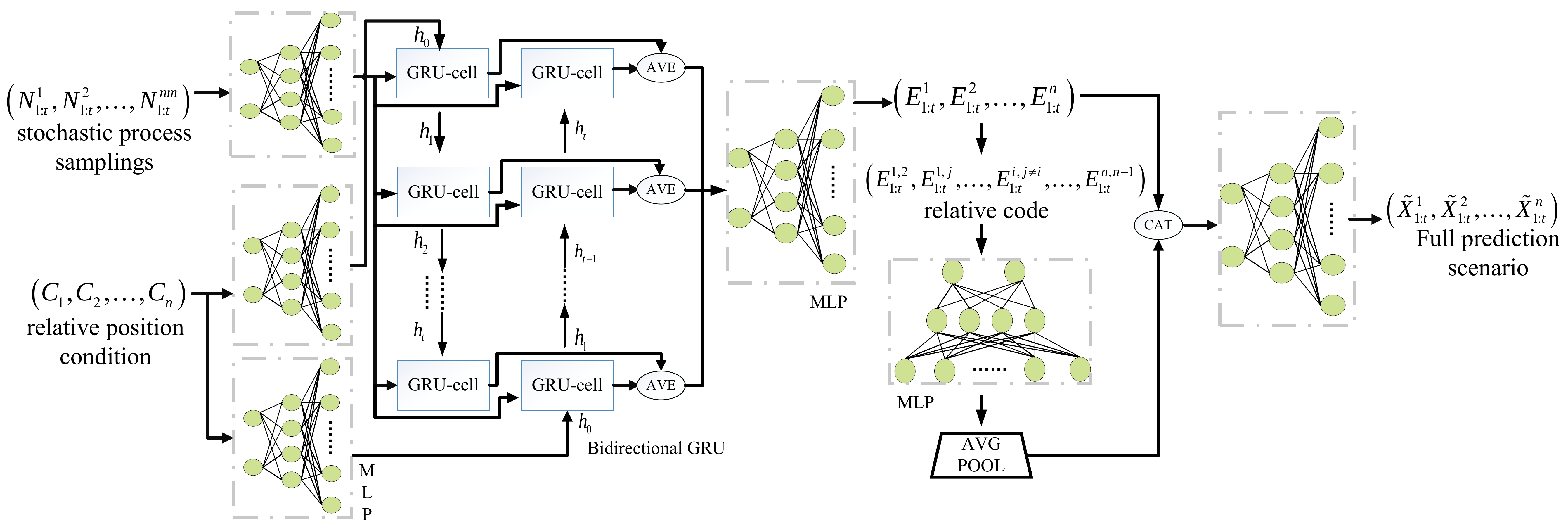}		
	\caption{Proposed generator framework. }
\end{figure*}

\subsection{MLP Based Regression Discriminator}

As generated prediction scenario data are used for vehicle trajectory prediction task, it is of significant importance to maintain sequential dependences in generated samples. For this, a regression discriminator is first proposed by us to distinguish multiple agents' real data from generated one. The regression discriminator learns to model joint distribution of inputs and outputs in a prediction task. Specifically, distribution of target vehicle's history data, target vehicle's future data and neighboring vehicles' history data are taken as inputs of the regression discriminator and modelled jointly. The regression discriminator outputs a classification probability that indicates degree of true.

Architecture of a regression discriminator is shown in \figurename{3}. For vehicle $j=1,2, \ldots, n$, trajectory data $\tilde X_{1: t}^{j}$ are pre-processed through
\begin{equation}
\begin{aligned}
\label{eqn_preprocess}
\hat{X}_{1: t}^{j} &= \tilde X_{1: t}^{j}-\tilde X_{t_h}^{e},\\
{X}_{1: t}^{j}&=\hat{X}_{1: t}^{j} /\left\|\hat X_{1: t}^{1}, \ldots, \hat X_{1: t}^{n}\right\|_{\infty}.
\end{aligned}
\end{equation}
After centering and normalisation pre-processing, target vehicle data are separated from neighboring agents' data. A MLP is applied to the target vehicle to encode its history $X_{1: t_{h}}^{e}$ and future data $X_{t_{h}+1: t_{h}+t_{f}}^{e}$ into $E_{r d}^{h}$ and $E_{r d}^{f}$ specifically. Relative difference between the target vehicle and neighboring vehicles are calculated and encoded into $E_{r d}^{n e i}$ through a MLP with a mean pooling layer, which is invariant to neighboring vehicles’ sequence. All codes are concatenated and encoded by two FC layers to get a feature vector $F_{rd}$. Finally, another FC layer is applied to the feature vector to obtain classification probalility $L_{rd}$.

Computation workflow can be formalized as:
\begin{equation}
\begin{aligned}
E_{r d}^{h}&=M_{r d 0}\left(X_{1: t_{h}}^{e}\right), \\
E_{r d}^{f}&=M_{r d 0}\left(X_{t_{h}+1: t_{h}+t_{f}}^{e}\right), \\
E_{r d}^{n e i}&=\left(\sum\nolimits_{i=1}^{n} M_{r d 1}\left(X_{1: t_{h}}^{e, i \neq e}\right)\right)/\left(n-1\right), \\
F_{r d}&=F C_{r d 0}\left(E_{r d}^{h}, E_{r d}^{f}, E_{r d}^{n e i}\right), \\
L_{r d}&=F C_{r d 1}\left(F_{r d}\right).
\end{aligned}
\end{equation}

\begin{figure}
	\centering
	\includegraphics[width=3.0in]{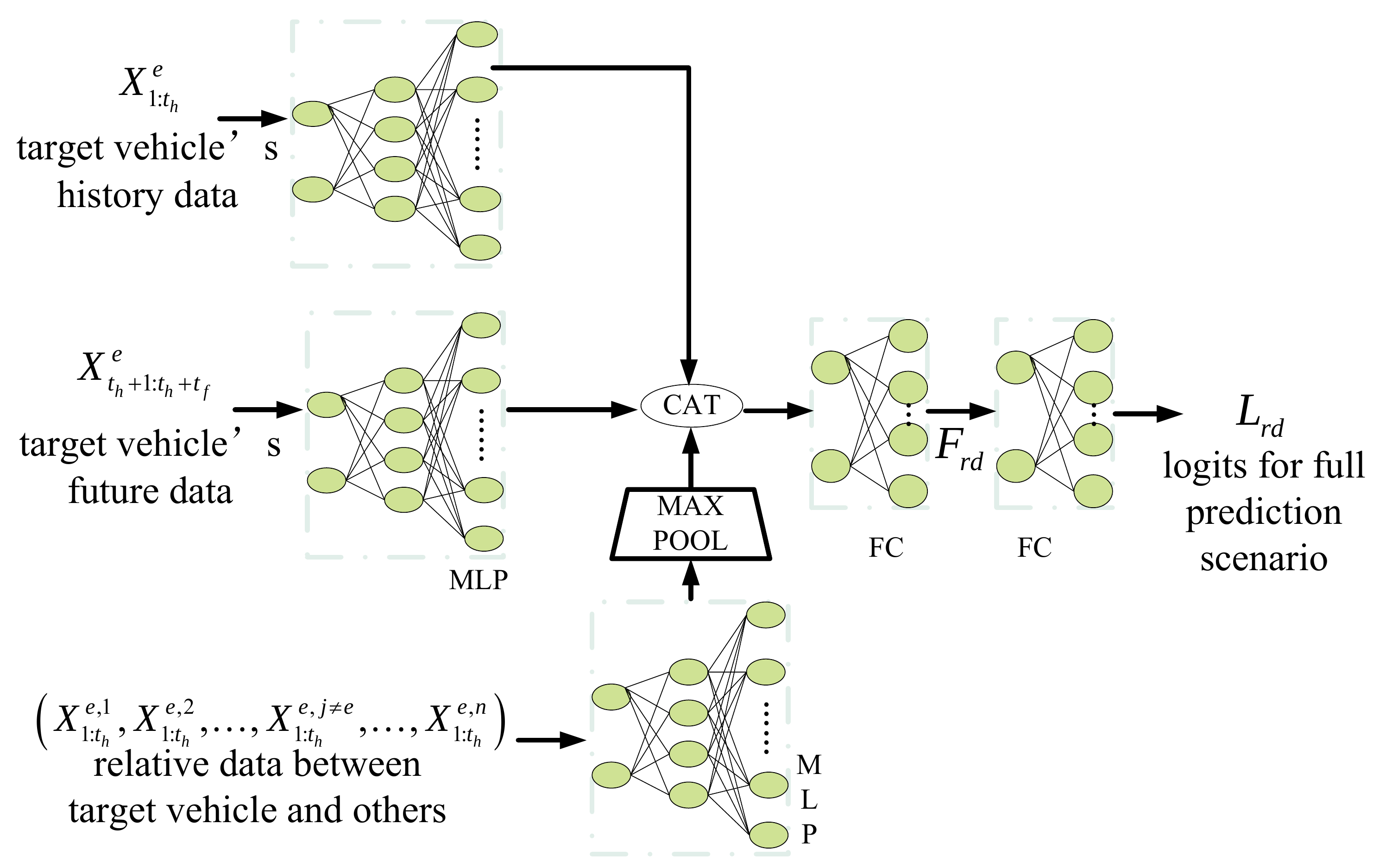}		
	\caption{Proposed regression discriminator framework. }
\end{figure}

\subsection{Evolution of Generation Model}
In a lifelong task chain $D=\{d_1,\dots,d_n\}$, generation models are required to be merged to a long-term model when a new task $d_{i+1}$ arrives. In general, there are two fusion methods. As illustrated in \figurename{4}, one method\cite{shin2017continual,lesort2019generative} merges generation model $G_i$ trained by task $d_i$ with the new task $d_{i+1}$. Generated samples from long-term model $G_i$ and real samples drawn from $d_{i+1}$ are combined as real samples to train a new generation model $G_{i+1}$, which we call Longterm-Data-Merge(LDM) method. Another method\cite{kamra2017deep,su2019generative} trains a temporal generation model $G_{i+1}^{t}$ for the new task $d_{i+1}$. Generated samples from long-term model $G_i$ and temporal model $G_{i+1}^{t}$ are combined as real samples to train a new long-term model $G_{i+1}$, which we call Longterm-Temporal-Merge(LTM) method. Although LDM performas better than LTM method intuitively, they are both applied to our lifelong task and compared.
\begin{figure*}[!t]
\centering
\subfloat[LDM]{\includegraphics[width=3.0in]{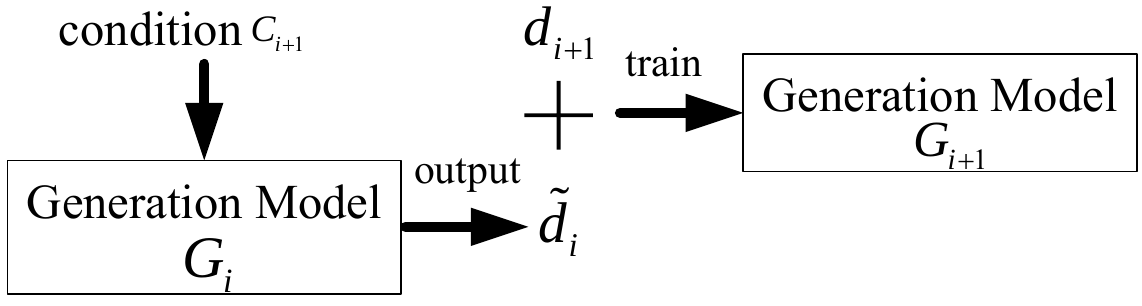}
\label{a}}
\hfil
\subfloat[LTM]{\includegraphics[width=3.0in]{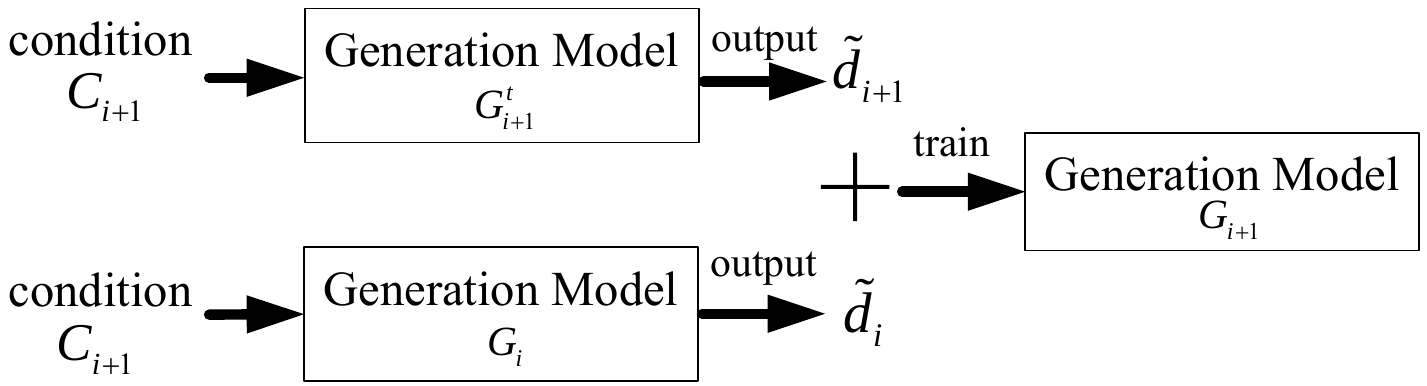}
\label{b}}
\caption{Two methods of generation model evolving.}
\label{LDM&LTM}
\end{figure*}

\subsection{Task model for vehicle trajectory prediction}
To perform vehicle trajectory prediction task with generated prediction scenario, a prediction model is proposed. As with mainstream trajectory prediction methods, history information and interactive relationship between target vehicle and neighboring vehicles are utilized to predict future trajectory of the target vehicle. Overall architecture is shown in \figurename{5}. First, target vehicle’s trajectory $X_{1: t_{h}}^{e}$ is separated from neighboring vehicles and encoded by a LSTM layer. Then, difference between target vehicle and others are encoded by a MLP and a mean pooling layer. These two parts are concatenated and encoded by a MLP further. A LSTM layer and a MLP is used to output predicted trajectories $\tilde{X}_{t_{h}+1: t_{h}+t_{f}}^{e}$.

Let $LSTM\left(\bullet\right)$ represents a LSTM unit, then for a prediction workflow, we have
\begin{equation}
\begin{aligned}
E_{r}^{h}&=M_{r 2}\left(L S T M\left(M_{r 0}\left(X_{1: t_{h}}^{e}\right)\right)\right), \\
E_{r d}^{n e i}&=M_{r 2}\left(\sum\nolimits_{i=1}^{n} M_{r 1}\left(X_{1: t_{h}}^{e, i \neq e}\right)\right), \\
\tilde{X}_{t_{h}+1: t_{h}+t_{f}}^{e}&=M_{r 3}\left(L S T M\left(E_{r}^{h}, E_{r d}^{n e i}\right)\right).
\end{aligned}
\end{equation}

\begin{figure*}
	\centering
	\includegraphics[width=6.0in]{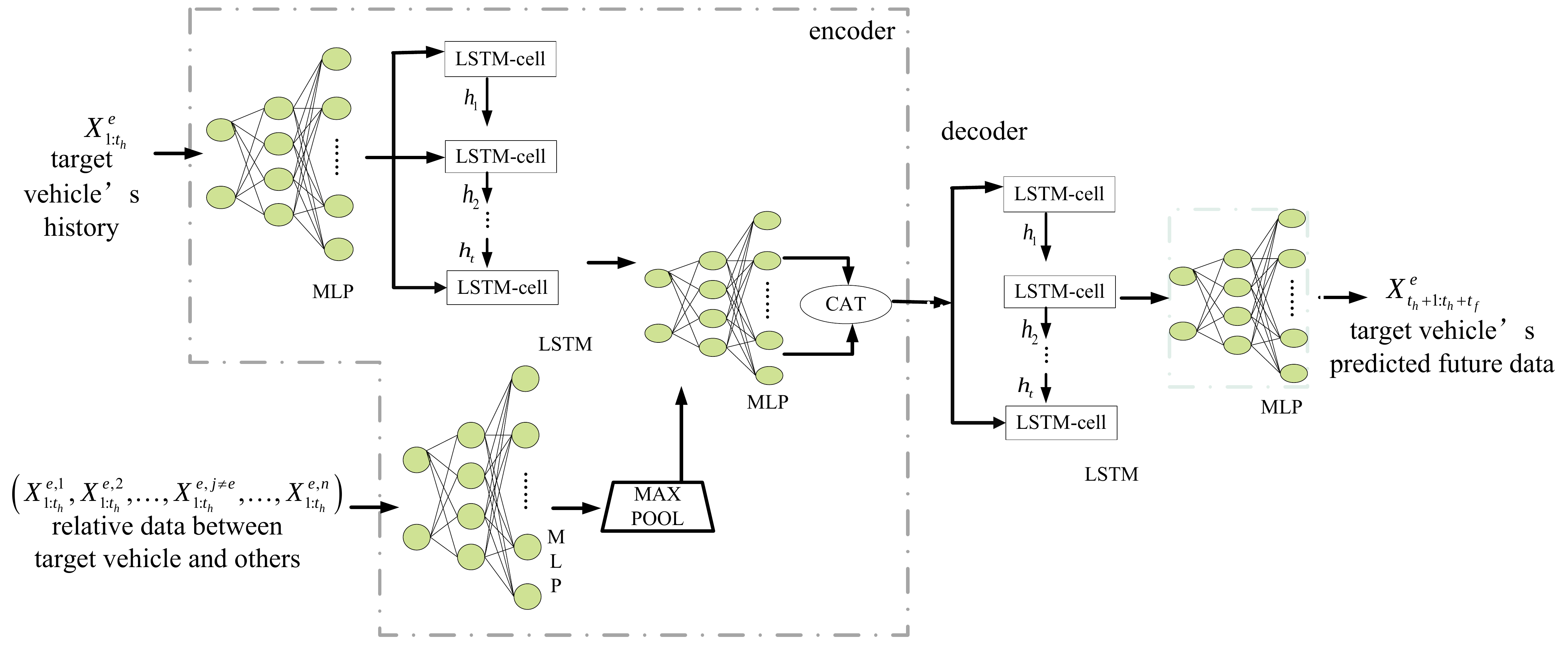}		
	\caption{Proposed vehicle trajectory prediction model. }
\end{figure*}

\section{Experiments and Analyses}
All experiments are realized via Pytorch\cite{paszke2019pytorch}. Running environment is Ubuntu 16.04, Intel Core i9-9900X CPU, GeForce GTX 1080 Ti, and 64GB RAM. All code including CKLD calculation and lifelong learning experiments and pre-processed data are available online$\footnote{https://github.com/CliffBao/GRTP}$. 
\subsection{Dataset and generation model setup}
To construct a lifelong task chain, five sub-datasets recorded in different locations are selected from four open datasets. Some traffic circumstances are illustrated in \figurename{6}.
\begin{figure*}
	\centering
	\includegraphics[width=6.0in]{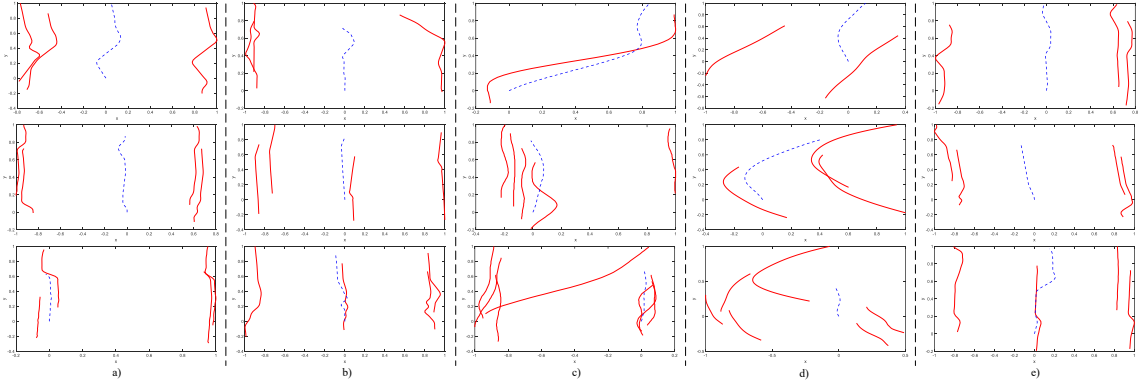}		
	\caption{Illustration of traffic circumstances in a) $US101\text{-}1$, b) $i801$, c) $highd20$, d) $inter5d$, and e) $US101\text{-}2$. Blue dash line indicates target vehicle. Red lines indicate neighboring vehicles. }
\end{figure*}
\begin{itemize}
	
\item 
NGSIM dataset\cite{alexiadis2004next}. 
The NGSIM dataset contains two sub-datasets, US101 dataset and I801 
dataset that are recorded on southbound US 101 and eastbound I-80 specifically. As tremendous vehicle trajectories are recorded, it’s time consuming to learn them all. Therefore, 7:50 a.m. to 8:05 a.m. trajectory records in US101 dataset are selected and named with $US101\text{-}1$. To simulate a case happened in real application where drives visit same place at 
different time period, 8:05 a.m. to 8:15 a.m trajectory records in US101 dataset are selected as $d_5$ and named with $US101\text{-}2$. Without losing generality, we keep full prediction scenarios that contains 4 and 5 vehicles only to ease the learning burden furtherly and over 188k items still remains. I801 dataset are refined as the same way and over 129k items remains. The selected datasets are separated into training, validation and testing dataset by 7:1:2.
\item HighD dataset\cite{highDdataset}. The highD dataset is a new dataset of naturalistic vehicle trajectories recorded at six different locations on Germany highways, which results in sixty recordings. Considering learning burden, the 20th recording is selected and pre-processed, which results in 88k items left for training. It is noted that as trajectories are recorded in 10HZ in NGSIM while 25HZ in HighD dataset, we aim to generate trajectories in 5HZ. As recording items is not comparable with NGSIM dataset in scale, full prediction scenarios that contains 2 to 9 vehicles are kept. For ease of representation, this dataset is called $highd20$.
\item Interaction dataset\cite{interactiondataset}. The interaction dataset contains naturalistic motions of various traffic participants in a variety of highly interactive driving scenarios from different countries. Trajectories in $\text{DR\_CHN\_Merging\_ZS}$ map is a lane merging dataset in China urban area. To be comparable with other datasets in scale, five trajectories records in the map are merged into a dataset named $inter5d$, which consists of 126k items.
\end{itemize}
Therefore, a lifelong task chain is formed by above five datasets
\begin{equation}
\begin{aligned}
D&=\{d_1,d_2,d_3,d_4,d_5\}\\
&=\{US101\text{-}1,i801,highd20,inter5d,US101\text{-}2\}.
\end{aligned}
\end{equation}

\subsection{Divergence between datasets}
To measure CKLD between two datasets, we fix maximum vehicle number to $n_v=5$ and only top $k=3$ eigen vectors are extracted. Gaussian hypothesis number of GMMs is set into $m=20$. The MDN is optimized through adam\cite{kingma2014adam} optimizer with learning rate $\gamma_m=0.0004$ and batch size $b_m=4096$. CKLD results of pairwise datasets are presented in Table I.
\begin{table}
\caption{CKLD between datasets}
\begin{tabular}{lccccc}
\hline
\diagbox{dataset 1}{CKLD}{dataset 2} & $d_1$ & $d_2$ & $d_3$ & $d_4$ & $d_5$ \\
\hline
$d_1$ & 0 & 22.14 & 503.22 & 98.43 & 16.72\\
$d_2$ & 29.00 & 0 & 756.03 & 98.57 & 24.44 \\
$d_3$ & 88.46 & 53.87 & 0 & 142.26 & 121.42 \\ 
$d_4$ & 75.05 & 63.54 & 652.00 & 0 & 62.50 \\
$d_5$ & 19.57 & 17.09 & 511.17 & 93.49 & 0 \\
\hline
\end{tabular}
\end{table}
In Table I, CKLD between $d_1$ and $d_5$ is the closest, which is expected as they are collected in the same highway at different time period. Divergence between $d_3$ and other datasets are quite large because only $d_3$ is colllected in Germany. The same goes for $d_4$.

\subsection{Validation of trajectory prediction model}
We compare proposed the task model performance with other benchmark trajectory prediction models.
\begin{itemize}\item Constant Velocity(CV). A constant velocity Kalman filter reported in \cite{deo2018convolutional}. \item GAIL-GRU. A generative adversarial imitation learning model described in \cite{kuefler2017imitating}. \item LSTM with fully connected social pooling (S-LSTM). This uses the fully connected social pooling described in \cite{alahi2016social} and generates a unimodal output distribution. \item LSTM with convolutional social pooling (CS-LSTM). This uses convolutional social pooling and generates a unimodal output distribution\cite{deo2018convolutional}. As designing an extrodinary prediction model is not our main goal, we use results reported in \cite{deo2018convolutional}.
\end{itemize}

\begin{table}
\begin{threeparttable}
\centering 
\caption{RMSE(m) comparison of proposed task model and other common methods on complete NGSIM dataset.}
\label{TAB1}
\begin{tabular}{cccccc}
\toprule
PH(s) &1 &2 &3 &4 &5 \\
\midrule
CV & 0.73 & 1.78 & 3.13 & 4.78 & 6.68 \\
GAIL-GRU & 0.69  & 1.51 & 2.55  & 3.65 & 4.71      \\ 
S-LSTM   &0.65   &1.31  &2.16  &3.25   &4.55      \\
CS-LSTM   &0.61   &\textbf{1.27}  &\textbf{2.09}   &\textbf{3.10}  &\textbf{4.37} \\
Ours   &\textbf{0.55}  &1.28  &2.18  &3.30  &4.64   \\
\bottomrule
\end{tabular}
\end{threeparttable}
\end{table}

To present a fair comparison, full NGSIM dataset is used to demonstrate validity of proposed prediction model, which is different from lifelong setups. RMSE performance for varied prediction horizon (PH) is given out in Table II. Suppose $BS$ batch size predicted future trajectories $\tilde{X}_{t}^{i}, i=1,2,\dots,BS$ at time $t$ are calculated and real future trajectories $X_{t}^{i}, i=1,2,\dots,BS$ are available, then RMSE at time $t$ is
\begin{equation}
RMSE\left(t\right)=\sqrt{\sum\nolimits_{i=1}^{BS}\left\|X_{t}^{i}-\tilde{X}_{t}^{i}\right\|^{2}/BS}.
\end{equation}
From Table II, it can be concluded that the task model possess similar prediction ability with mainstream methods. As we are not aiming to significantly improve prediction accuracy on single dataset but instead improve generality and lifelong prediction ability over multiple tasks, the performance of proposed prediction model may not compatible with state-of-the-art methods.

\subsection{Lifelong trajectory prediction}
In R2GAN, a generator takes inputs as several Gaussian process samplings and label conditions of trajectories that indicate relative position to a target vehicle. For example, -1 indicates a vehicle is located on the left lane of the target vehicle at the beginning. The generator outputs corresponding vehicle trajectory snippets lasting for 8 seconds, i.e. 41 steps. For a regression discriminator, real or generated prediction scenario are classified as real or fake. R2GAN is trained by adam\cite{kingma2014adam} optimizer with learning rate $\gamma=0.0001$. Non-saturating GAN loss function is applied as in\cite{goodfellow2016nips}.
To demonstrate lifelong prediction ability of proposed framework, three other methods are realized and compared with our approach.
\begin{itemize}
\item Generative replay based trajectory prediction(GRTP). This is the proposed lifelong trajectory prediction model based on generative replay. Resulted from two fusion methods LDM and LTM, the GRTP is furtherly classified into GRTP-D and GRTP-T specifically.
\item Joint training(JT). Joint training violates essential storage limitation and assumes that all data are available. This is regarded as the best possible performance over any lifelong learning methods. \item Fixed model(FM). A trajectory model trained by task $d_1$ will not be adjusted anymore and will be applied to new tasks directly. 
\item Fine tuning(FT). A trajectory model is trained while new task data $d_i$ are available. This is a possible choice but is expected to forget everything about old tasks. From some perspective, research works on trajctory prediction model design and optimization can be categoried into this method, although they did not test the model trained on new dataset on old ones.
\end{itemize} 
The RMSE plots through the full lifelong task chain is illustrated in \figurename{7}, where local RMSE around $t_f$ is zoomed in by a mini plot. As the lifelong task chain proceeds from $d_1$ to $d_5$, prediction performance on future 5 seconds horizon is validated on more tasks. Exact numeric result after addressing $d_5$ is given out in Table III. 
From experiment results, we can see that
\begin{itemize}
\item it is obvious FT forgets old knowledge while attaining new knowledge. Indeed, FT is a common practice when new task arrives and performs well if divergence between old and new task is small. From CKLD computation result, divergence between $d_1$, $d_2$, and $d_5$ is relatively small. Therefore, FT performs well on these three tasks after $d_5$. On the contrary, CKLD between $d_5$ and $d_3$, $d_4$ is relatively large, which results in poor performance on $d_3$ and $d_4$ after tuning on task $d_5$. The same consequence of applying FT to new task is also remarkable in \figurename{7}.
\item FM is trained on $d_1$ only. As a result, good performance on $d_1$ and $d_5$ is attainable after $d_5$ while large RMSE is observed on other tasks.
\item The proposed GRTP-T and GRTP-D perform well consistently over all tasks and possess close RMSE to JT. As JT stores all task data and can be considered as the best possible performance in lifelong task chain, we can conclude that GRTP mitigates catastrophic forgetting and realizes lifelong learning whether with LDM or LTM fusion method.
\item Although intuitively thinking, GRTP-D will outperforms GRTP-T as it merges longterm generation model with new data directly and avoids training a new generation model on new data, which avoids distribution learning bias introduced by the temporal generation model. However, it can be observed in Table III and \figurename{7} that no significant performance gap exists between them. This observation demonstrates that minor or even no distribution bias is introduced by our proposed R2GAN, which validates effectiveness of proposed R2GAN and model merging method implicitly.   
\end{itemize}

\begin{figure*}[!t]
	\centering
	\includegraphics[width=6.0in]{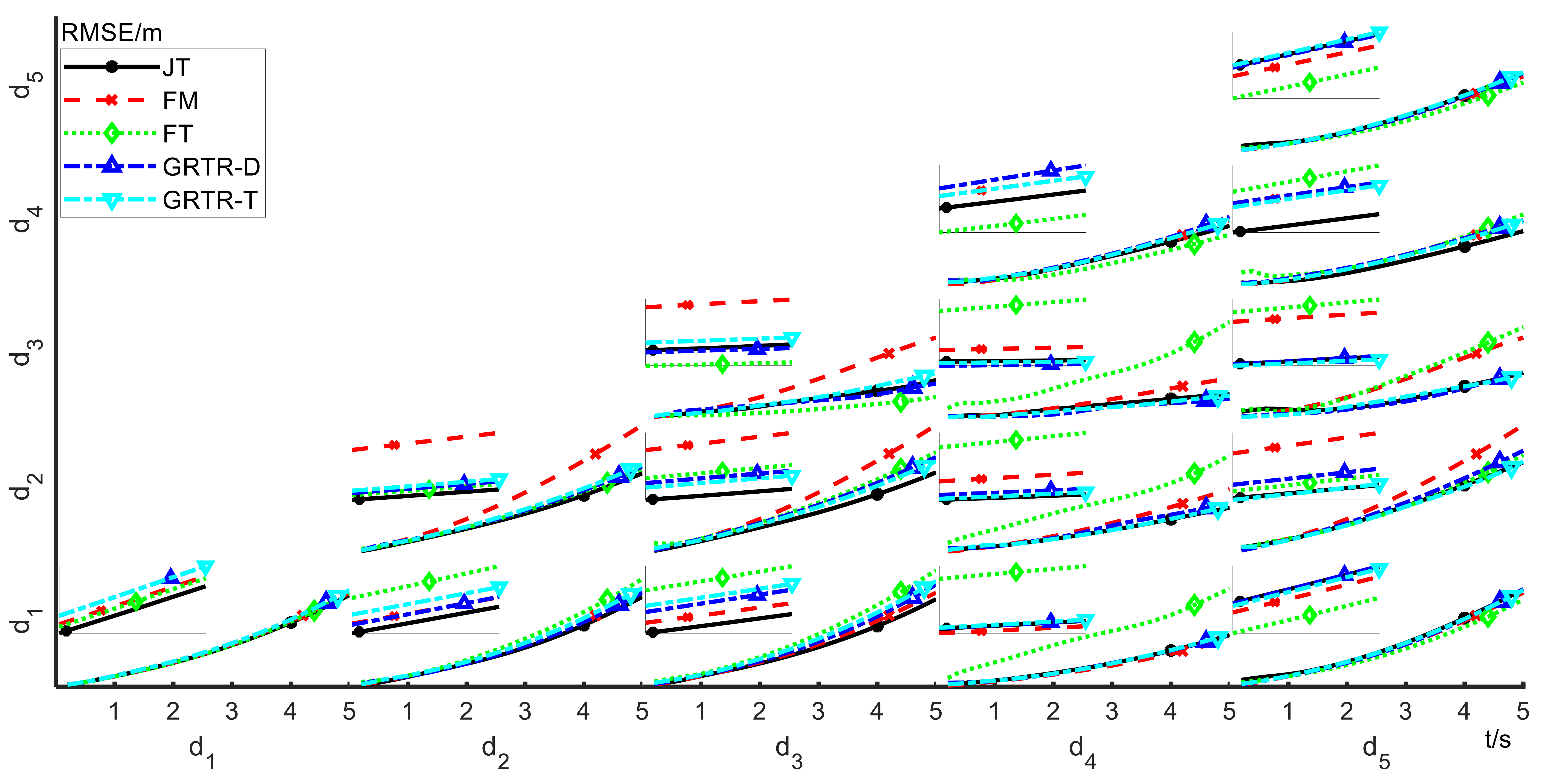}		
	\caption{RMSE for full lifelong task chain. }
\end{figure*}

\begin{table}
\begin{threeparttable}
\centering \caption{RMSE(m) comparison of different models after finishing lifelong task chain.}
\label{TAB2}
\begin{tabular}{ccccccc}
\toprule
 & PH(s) & 1 & 2 & 3 & 4 & 5\\
\midrule
\multirow{5}{*}{$d_1$} &JT & 0.93   & 1.86    &3.30    &5.14    &7.30 \\
                       &FM & 0.79   & 1.82    &3.18    &4.94    &7.06 \\
                       &FT & \textbf{0.79}   & \textbf{1.63}    &\textbf{2.86}    &\textbf{4.49}    &\textbf{6.53} \\
                       &GRTP-D & 0.80   & 1.80    &3.19    &5.06    &7.32 \\ 
                       &GRTP-T & 0.81   & 1.86    &3.29    &5.08    &7.25 \\\hline
\multirow{5}{*}{$d_2$} &JT & 1.03   & 2.15    &3.53    &\textbf{5.09}    &\textbf{6.80} \\
                       &FM & 1.08   & 2.56    &4.54    &6.91    &9.57 \\
                       &FT & 1.09   & \textbf{2.09}    &\textbf{3.48}    &5.26    &7.34 \\
                       &GRTP-D & 1.00   & 2.25    &3.79    &5.61    &7.67 \\ 
                       &GRTP-T & \textbf{0.99}   & 2.10    &3.49    &4.97    &6.85 \\\hline
\multirow{5}{*}{$d_3$} &JT & 1.46   & 1.66    &2.97    &4.94    & 7.01 \\
                       &FM & 1.36   & 3.30    &6.06    &9.24    &12.25 \\
                       &FT & 1.32   & 3.21    &6.49    &10.01   &13.81 \\
                       &GRTP-D & 0.81   &\textbf{1.47}    &\textbf{2.68}    &\textbf{4.66}    &7.07 \\ 
                       &GRTP-T & \textbf{0.70}   & 1.88    &3.41    &4.93    &\textbf{6.66} \\\hline
\multirow{5}{*}{$d_4$} &JT &\textbf{0.36}   &\textbf{0.96}    &\textbf{1.89}    &\textbf{2.95}    &\textbf{4.09} \\
                       &FM & 0.44   & 1.28    &2.34    &3.55    &4.86 \\
                       &FT & 0.79   & 1.32    &2.30    &3.69    &5.33 \\
                       &GRTP-D & 0.57   & 1.35    &2.33    &3.56    &4.89 \\ 
                       &GRTP-T & 0.47   & 1.25    &2.25    &3.46    &4.82 \\\hline
\multirow{5}{*}{$d_5$} &JT & 0.71   & 1.48    &2.70    &4.21    &5.92 \\
                       &FM & 0.65   & 1.52    &2.65    &4.07    &5.67 \\
                       &FT & 0.62   & \textbf{1.31}    &\textbf{2.32}    &\textbf{3.63}    &\textbf{5.20} \\
                       &GRTP-D &\textbf{0.62}   & 1.47    &2.64    &4.15    &5.90 \\ 
                       &GRTP-T & 0.63   & 1.54    &2.75    &4.24    &5.96 \\
\bottomrule
\end{tabular}
\end{threeparttable}
\end{table}

\section{Conclusion}
Maintaining consistent performance on vehicle trajectory prediction over different traffic circumstances is of significant importance for safe driving. Lifelong trajectory prediction is first addressed by us. Key problem hidering lifelong learning is catastrophic forgetting which arises from existence of spatiotemporal dependency divergence. To analysis the divergence between different traffic circumstances, CKLD is calculated based on GMMs approximation and MC sampling. Then a R2GAN is developed to generate dynamic number of vehicles in traffic circumstances, which guarantees inherent spatiotemporal dependency through a novel regression discriminator. Two methods are applied to construct a lifelong R2GAN model, LTM and LDM. LTM merges generated trajetories from long-term R2GAN and temporal R2GAN to train a new long-term generation model. LDM takes trajectories generated from long-term R2GAN and sampled from real dataset as real data to update the long-term generation model. Different spatiotemporal dependency are remembered by the long-term generation model that capable of generating samples from all processed tasks, thus mitigating catastrophic forgetting problem. Both merging method are validated through a constructed lifelong task chain and fulfill lifelong trajectory prediction task with consistent performance.  

In this work, four datasets are appplied to verify effectiveness of proposed lifelong framework. In future work, more diversified traffic circumstances can be introduced to improve lifelong prediction performance.

\section{Acknowledgment}

This work is supported by the National Natural Science Foundation of China (Grant No. 91848111, No. 62103393).

\bibliographystyle{IEEEtran}
\bibliography{IEEEabrv,lifelong}

\begin{thebibliography}{10}
\providecommand{\url}[1]{#1}
\csname url@samestyle\endcsname
\providecommand{\newblock}{\relax}
\providecommand{\bibinfo}[2]{#2}
\providecommand{\BIBentrySTDinterwordspacing}{\spaceskip=0pt\relax}
\providecommand{\BIBentryALTinterwordstretchfactor}{4}
\providecommand{\BIBentryALTinterwordspacing}{\spaceskip=\fontdimen2\font plus
\BIBentryALTinterwordstretchfactor\fontdimen3\font minus
  \fontdimen4\font\relax}
\providecommand{\BIBforeignlanguage}[2]{{%
\expandafter\ifx\csname l@#1\endcsname\relax
\typeout{** WARNING: IEEEtran.bst: No hyphenation pattern has been}%
\typeout{** loaded for the language `#1'. Using the pattern for}%
\typeout{** the default language instead.}%
\else
\language=\csname l@#1\endcsname
\fi
#2}}
\providecommand{\BIBdecl}{\relax}
\BIBdecl

\bibitem{2017Predicting}
S.~Qiao, N.~Han, J.~Wang, R.~H. Li, L.~A. Gutierrez, and X.~Wu, ``Predicting
  long-term trajectories of connected vehicles via the prefix-projection
  technique,'' \emph{IEEE Transactions on Intelligent Transportation Systems},
  pp. 1--11, 2017.

\bibitem{2019Interactive}
L.~Hou, L.~Xin, S.~E. Li, B.~Cheng, and W.~Wang, ``Interactive trajectory
  prediction of surrounding road users for autonomous driving using
  structural-lstm network,'' \emph{IEEE Transactions on Intelligent
  Transportation Systems}, vol.~PP, no.~99, pp. 1--11, 2019.

\bibitem{li2019interaction}
J.~Li, H.~Ma, and M.~Tomizuka, ``Interaction-aware multi-agent tracking and
  probabilistic behavior prediction via adversarial learning,'' in \emph{2019
  International Conference on Robotics and Automation (ICRA)}.\hskip 1em plus
  0.5em minus 0.4em\relax IEEE, 2019, pp. 6658--6664.

\bibitem{li2019grip}
X.~Li, X.~Ying, and M.~C. Chuah, ``Grip: Graph-based interaction-aware
  trajectory prediction,'' in \emph{2019 IEEE Intelligent Transportation
  Systems Conference (ITSC)}.\hskip 1em plus 0.5em minus 0.4em\relax IEEE,
  2019, pp. 3960--3966.

\bibitem{chandra2019traphic}
R.~Chandra, U.~Bhattacharya, A.~Bera, and D.~Manocha, ``Traphic: Trajectory
  prediction in dense and heterogeneous traffic using weighted interactions,''
  in \emph{Proceedings of the IEEE/CVF Conference on Computer Vision and
  Pattern Recognition}, 2019, pp. 8483--8492.

\bibitem{dulian2021multi}
A.~Dulian and J.~C. Murray, ``Multi-modal anticipation of stochastic
  trajectories in a dynamic environment with conditional variational
  autoencoders,'' \emph{arXiv preprint arXiv:2103.03912}, 2021.

\bibitem{bahari2021injecting}
M.~Bahari, I.~Nejjar, and A.~Alahi, ``Injecting knowledge in data-driven
  vehicle trajectory predictors,'' \emph{arXiv preprint arXiv:2103.04854},
  2021.

\bibitem{zyner2019naturalistic}
A.~Zyner, S.~Worrall, and E.~Nebot, ``Naturalistic driver intention and path
  prediction using recurrent neural networks,'' \emph{IEEE transactions on
  intelligent transportation systems}, vol.~21, no.~4, pp. 1584--1594, 2019.

\bibitem{ammoun2009real}
S.~Ammoun and F.~Nashashibi, ``Real time trajectory prediction for collision
  risk estimation between vehicles,'' in \emph{2009 IEEE 5th International
  Conference on Intelligent Computer Communication and Processing}.\hskip 1em
  plus 0.5em minus 0.4em\relax IEEE, 2009, pp. 417--422.

\bibitem{lytrivis2008cooperative}
P.~Lytrivis, G.~Thomaidis, and A.~Amditis, ``Cooperative path prediction in
  vehicular environments,'' in \emph{2008 11th International IEEE Conference on
  Intelligent Transportation Systems}.\hskip 1em plus 0.5em minus 0.4em\relax
  IEEE, 2008, pp. 803--808.

\bibitem{tran2014online}
Q.~Tran and J.~Firl, ``Online maneuver recognition and multimodal trajectory
  prediction for intersection assistance using non-parametric regression,'' in
  \emph{2014 IEEE Intelligent Vehicles Symposium Proceedings}.\hskip 1em plus
  0.5em minus 0.4em\relax IEEE, 2014, pp. 918--923.

\bibitem{aoude2011behavior}
G.~S. Aoude, V.~R. Desaraju, L.~H. Stephens, and J.~P. How, ``Behavior
  classification algorithms at intersections and validation using naturalistic
  data,'' in \emph{2011 IEEE Intelligent Vehicles Symposium (IV)}.\hskip 1em
  plus 0.5em minus 0.4em\relax IEEE, 2011, pp. 601--606.

\bibitem{kumar2013learning}
P.~Kumar, M.~Perrollaz, S.~Lefevre, and C.~Laugier, ``Learning-based approach
  for online lane change intention prediction,'' in \emph{2013 IEEE Intelligent
  Vehicles Symposium (IV)}.\hskip 1em plus 0.5em minus 0.4em\relax IEEE, 2013,
  pp. 797--802.

\bibitem{streubel2014prediction}
T.~Streubel and K.~H. Hoffmann, ``Prediction of driver intended path at
  intersections,'' in \emph{2014 IEEE Intelligent Vehicles Symposium
  Proceedings}.\hskip 1em plus 0.5em minus 0.4em\relax IEEE, 2014, pp.
  134--139.

\bibitem{xu2015establishing}
L.~Xu, J.~Hu, H.~Jiang, and W.~Meng, ``Establishing style-oriented driver
  models by imitating human driving behaviors,'' \emph{IEEE Transactions on
  Intelligent Transportation Systems}, vol.~16, no.~5, pp. 2522--2530, 2015.

\bibitem{deo2018would}
N.~Deo, A.~Rangesh, and M.~M. Trivedi, ``How would surround vehicles move? a
  unified framework for maneuver classification and motion prediction,''
  \emph{IEEE Transactions on Intelligent Vehicles}, vol.~3, no.~2, pp.
  129--140, 2018.

\bibitem{deo2018convolutional}
N.~Deo and M.~M. Trivedi, ``Convolutional social pooling for vehicle trajectory
  prediction,'' in \emph{Proceedings of the IEEE Conference on Computer Vision
  and Pattern Recognition Workshops}, 2018, pp. 1468--1476.

\bibitem{messaoud2019relational}
K.~Messaoud, I.~Yahiaoui, A.~Verroust-Blondet, and F.~Nashashibi, ``Relational
  recurrent neural networks for vehicle trajectory prediction,'' in \emph{2019
  IEEE Intelligent Transportation Systems Conference (ITSC)}.\hskip 1em plus
  0.5em minus 0.4em\relax IEEE, 2019, pp. 1813--1818.

\bibitem{gupta2018social}
A.~Gupta, J.~Johnson, L.~Fei-Fei, S.~Savarese, and A.~Alahi, ``Social gan:
  Socially acceptable trajectories with generative adversarial networks,'' in
  \emph{Proceedings of the IEEE Conference on Computer Vision and Pattern
  Recognition}, 2018, pp. 2255--2264.

\bibitem{sanderson1980pattern}
A.~C. Sanderson and A.~K. Wong, ``Pattern trajectory analysis of nonstationary
  multivariate data,'' \emph{IEEE Transactions on Systems, Man, and
  Cybernetics}, vol.~10, no.~7, pp. 384--392, 1980.

\bibitem{muller2007dynamic}
M.~M{\"u}ller, ``Dynamic time warping,'' \emph{Information retrieval for music
  and motion}, pp. 69--84, 2007.

\bibitem{robinson1990temporal}
M.~T. Robinson, ``The temporal development of collision cascades in the
  binary-collision approximation,'' \emph{Nuclear Instruments and Methods in
  Physics Research Section B: Beam Interactions with Materials and Atoms},
  vol.~48, no. 1-4, pp. 408--413, 1990.

\bibitem{ismail2015new}
A.~Ismail and A.~Vigneron, ``A new trajectory similarity measure for gps
  data,'' in \emph{Proceedings of the 6th ACM SIGSPATIAL International Workshop
  on GeoStreaming}, 2015, pp. 19--22.

\bibitem{pelekis2007similarity}
N.~Pelekis, I.~Kopanakis, G.~Marketos, I.~Ntoutsi, G.~Andrienko, and
  Y.~Theodoridis, ``Similarity search in trajectory databases,'' in \emph{14th
  International Symposium on Temporal Representation and Reasoning
  (TIME'07)}.\hskip 1em plus 0.5em minus 0.4em\relax IEEE, 2007, pp. 129--140.

\bibitem{su2020survey}
H.~Su, S.~Liu, B.~Zheng, X.~Zhou, and K.~Zheng, ``A survey of trajectory
  distance measures and performance evaluation,'' \emph{The VLDB Journal},
  vol.~29, no.~1, pp. 3--32, 2020.

\bibitem{https://doi.org/10.1002/cpa.3160360204}
\BIBentryALTinterwordspacing
M.~D. Donsker and S.~R.~S. Varadhan, ``Asymptotic evaluation of certain markov
  process expectations for large time. iv,'' \emph{Communications on Pure and
  Applied Mathematics}, vol.~36, no.~2, pp. 183--212, 1983. [Online].
  Available:
  \url{https://onlinelibrary.wiley.com/doi/abs/10.1002/cpa.3160360204}
\BIBentrySTDinterwordspacing

\bibitem{belghazi2018mutual}
M.~I. Belghazi, A.~Baratin, S.~Rajeshwar, S.~Ozair, Y.~Bengio, A.~Courville,
  and D.~Hjelm, ``Mutual information neural estimation,'' in
  \emph{International Conference on Machine Learning}.\hskip 1em plus 0.5em
  minus 0.4em\relax PMLR, 2018, pp. 531--540.

\bibitem{wang2006nearest}
Q.~Wang, S.~R. Kulkarni, and S.~Verd{\'u}, ``A nearest-neighbor approach to
  estimating divergence between continuous random vectors,'' in \emph{2006 IEEE
  International Symposium on Information Theory}.\hskip 1em plus 0.5em minus
  0.4em\relax IEEE, 2006, pp. 242--246.

\bibitem{gretton2006kernel}
A.~Gretton, K.~Borgwardt, M.~Rasch, B.~Sch{\"o}lkopf, and A.~Smola, ``A kernel
  method for the two-sample-problem,'' \emph{Advances in neural information
  processing systems}, vol.~19, pp. 513--520, 2006.

\bibitem{song2013kernel}
L.~Song, K.~Fukumizu, and A.~Gretton, ``Kernel embeddings of conditional
  distributions: A unified kernel framework for nonparametric inference in
  graphical models,'' \emph{IEEE Signal Processing Magazine}, vol.~30, no.~4,
  pp. 98--111, 2013.

\bibitem{ren2016conditional}
Y.~Ren, J.~Zhu, J.~Li, and Y.~Luo, ``Conditional generative moment-matching
  networks,'' \emph{Advances in Neural Information Processing Systems},
  vol.~29, pp. 2928--2936, 2016.

\bibitem{ren2019learning}
C.-X. Ren, P.~Ge, D.-Q. Dai, and H.~Yan, ``Learning kernel for conditional
  moment-matching discrepancy-based image classification,'' \emph{IEEE
  transactions on cybernetics}, 2019.

\bibitem{tabak2021data}
E.~G. Tabak, G.~Trigila, and W.~Zhao, ``Data driven conditional optimal
  transport,'' \emph{Machine Learning}, pp. 1--21, 2021.

\bibitem{cover1999elements}
T.~M. Cover, \emph{Elements of information theory}.\hskip 1em plus 0.5em minus
  0.4em\relax John Wiley \& Sons, 1999.

\bibitem{delange2021continual}
M.~Delange, R.~Aljundi, M.~Masana, S.~Parisot, X.~Jia, A.~Leonardis,
  G.~Slabaugh, and T.~Tuytelaars, ``A continual learning survey: Defying
  forgetting in classification tasks,'' \emph{IEEE Transactions on Pattern
  Analysis and Machine Intelligence}, 2021.

\bibitem{rusu2016progressive}
A.~A. Rusu, N.~C. Rabinowitz, G.~Desjardins, H.~Soyer, J.~Kirkpatrick,
  K.~Kavukcuoglu, R.~Pascanu, and R.~Hadsell, ``Progressive neural networks,''
  \emph{arXiv preprint arXiv:1606.04671}, 2016.

\bibitem{rosenfeld2018incremental}
A.~Rosenfeld and J.~K. Tsotsos, ``Incremental learning through deep
  adaptation,'' \emph{IEEE transactions on pattern analysis and machine
  intelligence}, vol.~42, no.~3, pp. 651--663, 2018.

\bibitem{lomonaco2017core50}
V.~Lomonaco and D.~Maltoni, ``Core50: a new dataset and benchmark for
  continuous object recognition,'' in \emph{Conference on Robot
  Learning}.\hskip 1em plus 0.5em minus 0.4em\relax PMLR, 2017, pp. 17--26.

\bibitem{li2017learning}
Z.~Li and D.~Hoiem, ``Learning without forgetting,'' \emph{IEEE transactions on
  pattern analysis and machine intelligence}, vol.~40, no.~12, pp. 2935--2947,
  2017.

\bibitem{kirkpatrick2017overcoming}
J.~Kirkpatrick, R.~Pascanu, N.~Rabinowitz, J.~Veness, G.~Desjardins, A.~A.
  Rusu, K.~Milan, J.~Quan, T.~Ramalho, A.~Grabska-Barwinska \emph{et~al.},
  ``Overcoming catastrophic forgetting in neural networks,'' \emph{Proceedings
  of the national academy of sciences}, vol. 114, no.~13, pp. 3521--3526, 2017.

\bibitem{aljundi2017expert}
R.~Aljundi, P.~Chakravarty, and T.~Tuytelaars, ``Expert gate: Lifelong learning
  with a network of experts,'' in \emph{Proceedings of the IEEE Conference on
  Computer Vision and Pattern Recognition}, 2017, pp. 3366--3375.

\bibitem{yao2019adversarial}
X.~Yao, T.~Huang, C.~Wu, R.-X. Zhang, and L.~Sun, ``Adversarial feature
  alignment: Avoid catastrophic forgetting in incremental task lifelong
  learning,'' \emph{Neural computation}, vol.~31, no.~11, pp. 2266--2291, 2019.

\bibitem{hou2018lifelong}
S.~Hou, X.~Pan, C.~C. Loy, Z.~Wang, and D.~Lin, ``Lifelong learning via
  progressive distillation and retrospection,'' in \emph{Proceedings of the
  European Conference on Computer Vision (ECCV)}, 2018, pp. 437--452.

\bibitem{rebuffi2017icarl}
S.-A. Rebuffi, A.~Kolesnikov, G.~Sperl, and C.~H. Lampert, ``icarl: Incremental
  classifier and representation learning,'' in \emph{Proceedings of the IEEE
  conference on Computer Vision and Pattern Recognition}, 2017, pp. 2001--2010.

\bibitem{lopez2017gradient}
D.~Lopez-Paz and M.~Ranzato, ``Gradient episodic memory for continual
  learning,'' \emph{arXiv preprint arXiv:1706.08840}, 2017.

\bibitem{shin2017continual}
H.~Shin, J.~K. Lee, J.~Kim, and J.~Kim, ``Continual learning with deep
  generative replay,'' \emph{arXiv preprint arXiv:1705.08690}, 2017.

\bibitem{su2019generative}
X.~Su, S.~Guo, T.~Tan, and F.~Chen, ``Generative memory for lifelong
  learning,'' \emph{IEEE transactions on neural networks and learning systems},
  vol.~31, no.~6, pp. 1884--1898, 2019.

\bibitem{lesort2019generative}
T.~Lesort, H.~Caselles-Dupr{\'e}, M.~Garcia-Ortiz, A.~Stoian, and D.~Filliat,
  ``Generative models from the perspective of continual learning,'' in
  \emph{2019 International Joint Conference on Neural Networks (IJCNN)}.\hskip
  1em plus 0.5em minus 0.4em\relax IEEE, 2019, pp. 1--8.

\bibitem{lesort2019marginal}
T.~Lesort, A.~Gepperth, A.~Stoian, and D.~Filliat, ``Marginal replay vs
  conditional replay for continual learning,'' in \emph{International
  Conference on Artificial Neural Networks}.\hskip 1em plus 0.5em minus
  0.4em\relax Springer, 2019, pp. 466--480.

\bibitem{liu2020generative}
X.~Liu, C.~Wu, M.~Menta, L.~Herranz, B.~Raducanu, A.~D. Bagdanov, S.~Jui, and
  J.~v. de~Weijer, ``Generative feature replay for class-incremental
  learning,'' in \emph{Proceedings of the IEEE/CVF Conference on Computer
  Vision and Pattern Recognition Workshops}, 2020, pp. 226--227.

\bibitem{li2020incremental}
H.~Li, W.~Dong, and B.-G. Hu, ``Incremental concept learning via online
  generative memory recall,'' \emph{IEEE Transactions on Neural Networks and
  Learning Systems}, 2020.

\bibitem{van2018generative}
G.~M. Van~de Ven and A.~S. Tolias, ``Generative replay with feedback
  connections as a general strategy for continual learning,'' \emph{arXiv
  preprint arXiv:1809.10635}, 2018.

\bibitem{kamra2017deep}
N.~Kamra, U.~Gupta, and Y.~Liu, ``Deep generative dual memory network for
  continual learning,'' \emph{arXiv preprint arXiv:1710.10368}, 2017.

\bibitem{wu2018memory}
C.~Wu, L.~Herranz, X.~Liu, Y.~Wang, J.~Van~de Weijer, and B.~Raducanu, ``Memory
  replay gans: learning to generate images from new categories without
  forgetting,'' \emph{arXiv preprint arXiv:1809.02058}, 2018.

\bibitem{xiang2019incremental}
Y.~Xiang, Y.~Fu, P.~Ji, and H.~Huang, ``Incremental learning using conditional
  adversarial networks,'' in \emph{Proceedings of the IEEE/CVF International
  Conference on Computer Vision}, 2019, pp. 6619--6628.

\bibitem{aljundi2018memory}
R.~Aljundi, F.~Babiloni, M.~Elhoseiny, M.~Rohrbach, and T.~Tuytelaars, ``Memory
  aware synapses: Learning what (not) to forget,'' in \emph{Proceedings of the
  European Conference on Computer Vision (ECCV)}, 2018, pp. 139--154.

\bibitem{fernando2017pathnet}
C.~Fernando, D.~Banarse, C.~Blundell, Y.~Zwols, D.~Ha, A.~A. Rusu, A.~Pritzel,
  and D.~Wierstra, ``Pathnet: Evolution channels gradient descent in super
  neural networks,'' \emph{arXiv preprint arXiv:1701.08734}, 2017.

\bibitem{lee2019joint}
D.~Lee, Y.~Gu, J.~Hoang, and M.~Marchetti-Bowick, ``Joint interaction and
  trajectory prediction for autonomous driving using graph neural networks,''
  \emph{arXiv preprint arXiv:1912.07882}, 2019.

\bibitem{chandra2020forecasting}
R.~Chandra, T.~Guan, S.~Panuganti, T.~Mittal, U.~Bhattacharya, A.~Bera, and
  D.~Manocha, ``Forecasting trajectory and behavior of road-agents using
  spectral clustering in graph-lstms,'' \emph{IEEE Robotics and Automation
  Letters}, vol.~5, no.~3, pp. 4882--4890, 2020.

\bibitem{bishop1994mixture}
C.~M. Bishop, ``Mixture density networks,'' 1994.

\bibitem{rothfuss2019conditional}
J.~Rothfuss, F.~Ferreira, S.~Walther, and M.~Ulrich, ``Conditional density
  estimation with neural networks: Best practices and benchmarks,'' \emph{arXiv
  preprint arXiv:1903.00954}, 2019.

\bibitem{hershey2007approximating}
J.~R. Hershey and P.~A. Olsen, ``Approximating the kullback leibler divergence
  between gaussian mixture models,'' in \emph{2007 IEEE International
  Conference on Acoustics, Speech and Signal Processing-ICASSP'07},
  vol.~4.\hskip 1em plus 0.5em minus 0.4em\relax IEEE, 2007, pp. IV--317.

\bibitem{BAO2022370}
\BIBentryALTinterwordspacing
P.~Bao, Z.~Chen, J.~Wang, and D.~Dai, ``Multiple agents’ spatiotemporal data
  generation based on recurrent regression dual discriminator gan,''
  \emph{Neurocomputing}, vol. 468, pp. 370--383, 2022. [Online]. Available:
  \url{https://www.sciencedirect.com/science/article/pii/S0925231221015344}
\BIBentrySTDinterwordspacing

\bibitem{goodfellow2014generative}
I.~J. Goodfellow, J.~Pouget-Abadie, M.~Mirza, B.~Xu, D.~Warde-Farley, S.~Ozair,
  A.~Courville, and Y.~Bengio, ``Generative adversarial networks,'' \emph{arXiv
  preprint arXiv:1406.2661}, 2014.

\bibitem{wiese2020quant}
M.~Wiese, R.~Knobloch, R.~Korn, and P.~Kretschmer, ``Quant gans: Deep
  generation of financial time series,'' \emph{Quantitative Finance}, vol.~20,
  no.~9, pp. 1419--1440, 2020.

\bibitem{karpathy2015deep}
A.~Karpathy and L.~Fei-Fei, ``Deep visual-semantic alignments for generating
  image descriptions,'' in \emph{Proceedings of the IEEE conference on computer
  vision and pattern recognition}, 2015, pp. 3128--3137.

\bibitem{vinyals2015show}
O.~Vinyals, A.~Toshev, S.~Bengio, and D.~Erhan, ``Show and tell: A neural image
  caption generator,'' in \emph{Proceedings of the IEEE conference on computer
  vision and pattern recognition}, 2015, pp. 3156--3164.

\bibitem{paszke2019pytorch}
A.~Paszke, S.~Gross, F.~Massa, A.~Lerer, J.~Bradbury, G.~Chanan, T.~Killeen,
  Z.~Lin, N.~Gimelshein, L.~Antiga \emph{et~al.}, ``Pytorch: An imperative
  style, high-performance deep learning library,'' \emph{Advances in neural
  information processing systems}, vol.~32, pp. 8026--8037, 2019.

\bibitem{alexiadis2004next}
V.~Alexiadis, J.~Colyar, J.~Halkias, R.~Hranac, and G.~McHale, ``The next
  generation simulation program,'' \emph{Institute of Transportation Engineers.
  ITE Journal}, vol.~74, no.~8, p.~22, 2004.

\bibitem{highDdataset}
R.~Krajewski, J.~Bock, L.~Kloeker, and L.~Eckstein, ``The highd dataset: A
  drone dataset of naturalistic vehicle trajectories on german highways for
  validation of highly automated driving systems,'' in \emph{2018 21st
  International Conference on Intelligent Transportation Systems (ITSC)}, 2018,
  pp. 2118--2125.

\bibitem{interactiondataset}
W.~Zhan, L.~Sun, D.~Wang, H.~Shi, A.~Clausse, M.~Naumann, J.~K\"ummerle,
  H.~K\"onigshof, C.~Stiller, A.~de~La~Fortelle, and M.~Tomizuka,
  ``{INTERACTION} {Dataset}: {An} {INTERnational}, {Adversarial} and
  {Cooperative} {moTION} {Dataset} in {Interactive} {Driving} {Scenarios} with
  {Semantic} {Maps},'' \emph{arXiv:1910.03088 [cs, eess]}, 2019.

\bibitem{kingma2014adam}
D.~P. Kingma and J.~Ba, ``Adam: A method for stochastic optimization,''
  \emph{arXiv preprint arXiv:1412.6980}, 2014.

\bibitem{kuefler2017imitating}
A.~Kuefler, J.~Morton, T.~Wheeler, and M.~Kochenderfer, ``Imitating driver
  behavior with generative adversarial networks,'' in \emph{2017 IEEE
  Intelligent Vehicles Symposium (IV)}.\hskip 1em plus 0.5em minus 0.4em\relax
  IEEE, 2017, pp. 204--211.

\bibitem{alahi2016social}
A.~Alahi, K.~Goel, V.~Ramanathan, A.~Robicquet, L.~Fei-Fei, and S.~Savarese,
  ``Social lstm: Human trajectory prediction in crowded spaces,'' in
  \emph{Proceedings of the IEEE conference on computer vision and pattern
  recognition}, 2016, pp. 961--971.

\bibitem{goodfellow2016nips}
I.~Goodfellow, ``Nips 2016 tutorial: Generative adversarial networks,''
  \emph{arXiv preprint arXiv:1701.00160}, 2016.

\end{thebibliography}
\begin{IEEEbiography}[{\includegraphics[width=1in,height=1.25in,clip,keepaspectratio]{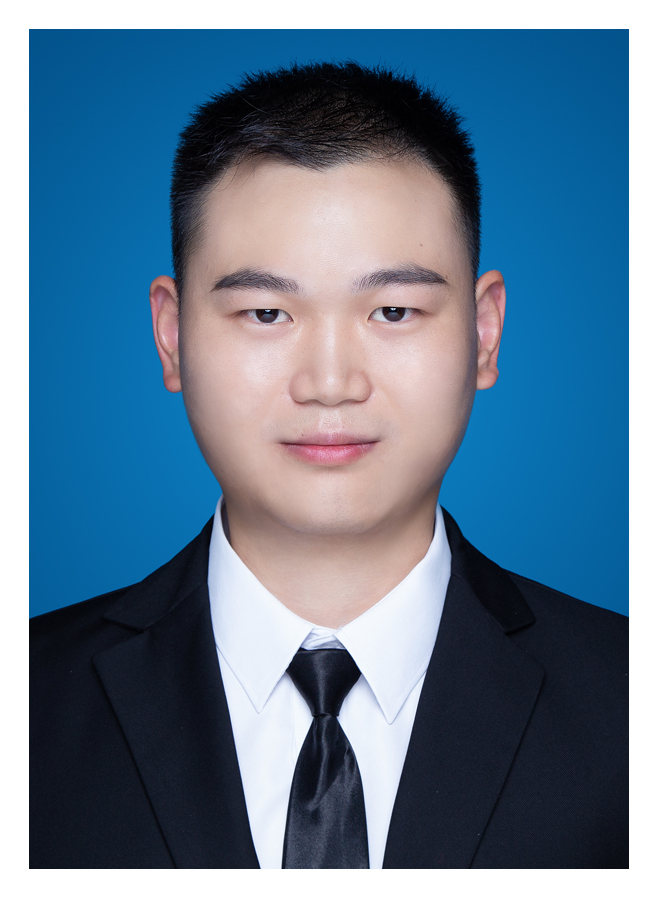}}]{Peng Bao}
Peng Bao received the B.S. degree from the University of Science and Technology of China (USTC) in 2014. He is now a PhD candidate in the Department of Automation, USTC. His research interests include intelligent information processing, mobile robots SLAM and deep learning.
\end{IEEEbiography}
\begin{IEEEbiography}[{\includegraphics[width=1in,height=1.25in,clip,keepaspectratio]{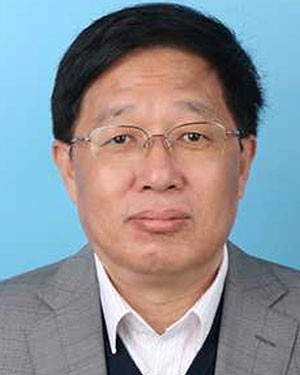}}]{Zonghai Chen}
Zonghai Chen was born in Anhui, China, in 1963. He received the B.S. degree in automation and the M.E. degree in control theory and control engineering from the University of Science and Technology of China (USTC), Hefei, China, in 1988 and 1991, respectively. He has been a Professor with the Department of Automation, USTC, since 1998. His research interests include modeling and control of complex systems, intelligent robotic and information processing, energy management technologies for electric vehicles, and smart microgrids. Prof. Chen is a member of the Robotics Technical Committee and Modelling, Identification and Signal Processing Technical Committee of the International Federation of Automation Control (IFAC). He was a recipient of special allowances from the State Council of PR China.
\end{IEEEbiography}
\begin{IEEEbiography}[{\includegraphics[width=1in,height=1.25in,clip,keepaspectratio]{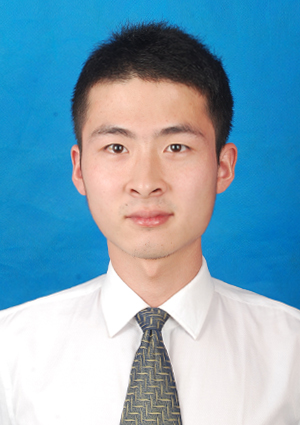}}]{Jikai Wang}
JiKai Wang received the B.S. degree from the University of Yanshan in 2014. He is now a post doctor in the Department of Automation, University of Science and Technology of China (USTC), China. His research interests include knowledge representation, intelligent information processing, robotics, visual SLAM, and machine learning.
\end{IEEEbiography}
\begin{IEEEbiography}[{\includegraphics[width=1in,height=1.25in,clip,keepaspectratio]{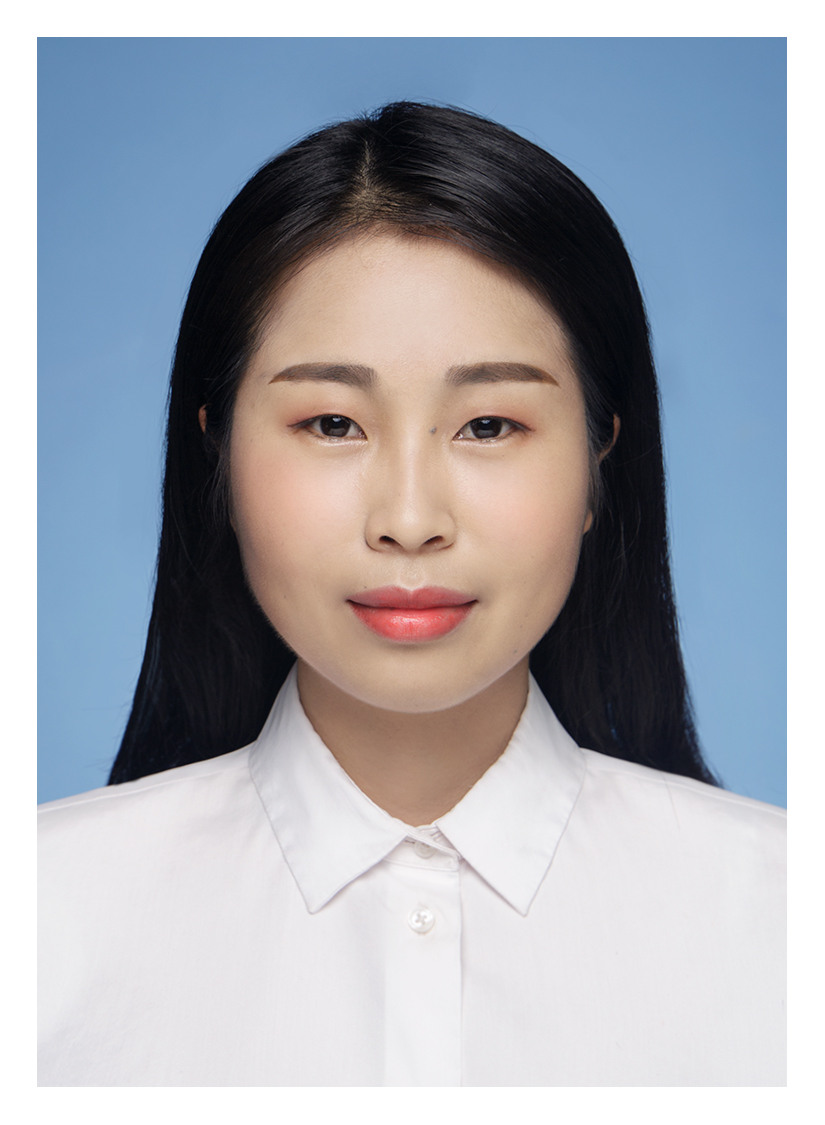}}]{Deyun Dai}
Deyun Dai received the B.S. degree from Harbin Engineering University, in 2016. She is now a PhD candidate in the Department of Automation, USTC. Her research interests include computer vision, environment perception in autonomous driving scenarios.
\end{IEEEbiography}
\begin{IEEEbiography}[{\includegraphics[width=1in,height=1.25in,clip,keepaspectratio]{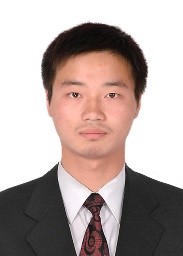}}]{Hao Zhao}
Hao Zhao received the B.S. and M.S. degree from the Southwest University of Science and Technology in 2014 and 2017. He is now a PhD candidate in the Department of Automation, University of Science and Technology of China (USTC). His research interests include object recognition and detection, scene perception and knowledge representation.

\end{IEEEbiography}

\end{document}